\documentclass[letterpaper, 10 pt, conference]{ieeeconf}  
\IEEEoverridecommandlockouts

\usepackage{algpseudocode}
\usepackage{algorithm}
\usepackage{graphicx} 
\usepackage{amsmath} 
\usepackage{amssymb} 
\title{\LARGE \bf
HSFN: Hierarchical Selection for Fake News Detection building Heterogeneous Ensemble
}

\author{Sara B. Coutinho$^{1}$, Rafael M.O. Cruz $^{2}$, Francimaria R. S. Nascimento$^{1}$ and George D. C. Cavalcanti$^{1}$
\thanks{This work was supported by the Brazilian CAPES (Coordenação de
 Aperfeiçoamento de Pessoal de Nível Superior, in portuguese).}
\thanks{$^{1}$Sara B. Coutinho, Francimaria R. S. Nascimento, George D. C. Cavalcanti are with Centro de Informática (CIn), Universidade Federal de Pernambuco (UFPE), Av. Jornalista Anibal Fernandes s/n, Recife, Brazil
        {\tt\small email: \{sbc2, frsn2, gdcc\}@cin.ufpe.br}}
\thanks{$^{2}$Rafael M. O. Cruz is with École de Technologie Supérieure (ÉTS), Université du Québec, 1100 Notre-Dame St W, Montreal, Quebec, 
Canada
        {\tt\small email: rafael.menelau-cruz@etsmtl.ca}}
}

\begin{document}

\maketitle
\thispagestyle{empty}
\pagestyle{empty}

\begin{abstract}

Psychological biases, such as confirmation bias, make individuals particularly vulnerable to believing and spreading fake news on social media, leading to significant consequences in domains such as public health and politics. Machine learning–based fact-checking systems have been widely studied to mitigate this problem. Among them, ensemble methods are particularly effective in combining multiple classifiers to improve robustness. However, their performance heavily depends on the diversity of the constituent classifiers—selecting genuinely diverse models remains a key challenge, especially when models tend to learn redundant patterns. In this work, we propose a novel automatic classifier selection approach that prioritizes diversity, also extended by performance. The method first computes pairwise diversity between classifiers and applies hierarchical clustering to organize them into groups at different levels of granularity. A HierarchySelect then explores these hierarchical levels to select one pool of classifiers per level, each representing a distinct intra-pool diversity. The most diverse pool is identified and selected for ensemble construction from these. The selection process incorporates an evaluation metric reflecting each classifier’s performance to ensure the ensemble also generalises well. We conduct experiments with 40 heterogeneous classifiers across six datasets from different application domains and with varying numbers of classes. Our method is compared against the Elbow heuristic and state-of-the-art baselines. Results show that our approach achieves the highest accuracy on two of six datasets. The implementation details are available on the project's repository: https://github.com/SaraBCoutinho/HSFN.

\end{abstract}

\section{INTRODUCTION}

The widespread adoption of the Internet, the emergence of new communication channels, and the advancement of artificial intelligence technologies have significantly improved access to information, offering various benefits to society. However, these same developments have facilitated the rapid dissemination of Fake News (FN), raising serious concerns as AI-driven algorithms amplify sensational content to maximize user engagement~\cite{bontridder2021role}. 
Several societal domains, including politics and public health, have been negatively affected. For instance, according to \cite{b1}, the 2016 U.S. presidential election may have been influenced by the spread of FN, as individuals tend to perceive information that aligns with their beliefs as accurate—a phenomenon rooted in political bias. Similarly, during the COVID-19 pandemic, FN related to the virus spread widely, contributing to public misinformation during a critical global crisis \cite{b3}. These impacts demonstrate the importance of developing effective strategies to detect and prevent FN dissemination.

Numerous studies have proposed machine learning-based fact-checking systems to address FN detection \cite{shu2017fake}. Among these, ensemble learning has emerged as a promising approach by combining the predictive power of multiple classifiers. As shown in \cite{farhangian2024fake}, integrating diverse feature representations and learning algorithms can enhance the performance of such systems. Nevertheless, ensuring classifier diversity remains a challenge, as redundant models can limit the ensemble's effectiveness. In fact, recent work demonstrates that combining all models leads to inferior results~\cite{pereira2025multiviewautoencodersfakenews}. Redundancy among classifiers may result in correlated errors, reducing the ensemble's ability to generalize and detect novel patterns in the data. To address this, \cite{b7} highlights the importance of analyzing the information dissimilarity between classifiers. In this context, \cite{b10} introduces a Multiple Classifier System (MCS) that relies on manual selection of classifiers from visually identified groups in a two-dimensional space called the Classifier Projection Space~\cite{pkekalska2002discussion}. While effective, this manual approach is prone to subjective bias and may fail to identify the most diverse combination of classifiers, reinforcing the need for automated selection strategies.

To address the challenges associated with constructing effective Multiple Classifier Systems (MCS) for FN detection, this work introduces the \textbf{Hierarchical Selection for Fake News Detection} (\textbf{HSFN}) framework. HSFN automatically builds ensembles by combining heterogeneous classifiers generated from diverse feature representation techniques and classification algorithms. During the training phase, a pool of classifiers is created, and their prediction behaviors on a validation set are analyzed using a diversity measure to compute a dissimilarity matrix. A hierarchical clustering process is then applied to group classifiers based on their diversity, forming a dendrogram that organizes classifiers according to their similarities. To select an effective and diverse subset, the proposed HSFN systematically explores different levels of the dendrogram, partitioning it into clusters and selecting representative classifiers from each group. By modeling classifier diversity hierarchically and automating the selection process, HSFN constructs robust ensembles while avoiding manual selection strategies from previous works~\cite{b10,b44}.

Hence, the main contributions of this work are:
(1) a comprehensive evaluation of MCS for FN detection;
(2) an exploration of classifier dissimilarity at different levels of granularity to enhance ensemble diversity and performance;
(3) the development of an automated selection strategy to identify the most diverse classifier pairs, considering both feature representations and algorithms. 

\section{RELATED WORKS}
\label{section:related_work}

\subsection{Feature representation for FN detection}

The development of various feature representation techniques over time has significantly influenced the methods used in fake news (FN) detection. In \cite{b30}, the authors employed an attention-based representation using BERT \cite{devlin2019bert} to capture semantic relationships by considering all elements of a sentence. Similarly, in \cite{b41}, the researchers investigated both sparse and dense representations, including TF-IDF and W2V \cite{mikolov2013efficient}. These were used to model temporal features through Bi-LSTM and spatial features via Named-Entity Recognition (NER) and GloVe embeddings \cite{pennington2014glove}. The resulting feature vectors were then combined, leading to improved FN detection performance. However, each feature representation technique presents distinct strengths and limitations that must be considered when selecting an appropriate method.

It is important to account for the specific characteristics of each technique. According to \cite{b34}, sparse vector representations such as TF-IDF, which rely on word frequency, are less suitable for large corpora. In contrast, dense representations like W2V are more appropriate for longer documents as they better capture contextual information. Additionally, GloVe provides dense representations based on global co-occurrence statistics, offering an effective alternative for capturing semantic meaning. Attention-based representations, such as those generated by BERT, have demonstrated strong performance in capturing contextual word meanings and exhibit greater flexibility \cite{b34, b37}. However, they require substantial computational resources, larger amounts of data, and transfer learning with labeled data for specific downstream tasks.

\subsection{Learning algorithm for FN detection}

Several works have used multiple algorithms to learn patterns for FN detection. In \cite{b13}, the authors explored two hybrid systems approaches. The first adopted a classifier system composed of classical machine learning algorithms, SVM, and deep learning algorithms, CNN, and DNN. The second explored only deep learning algorithms, CNN, and Bi-LSTM. Alternatively, \cite{b14} used an ensemble system that combined the algorithms instead of joining them one after the other in a sequence to leverage collective learning simultaneously. In \cite{b31}, the researchers utilised a stacking model using SVM, LR, and RF classifier algorithms and XGBoost as meta-classifiers. 
\cite{b15} highlights the effectiveness of ensemble systems with heterogeneous algorithms, demonstrating that stacking outperforms hybrid systems that employ sequential algorithmic combinations.

\subsection{MCS for FN detection}

In \cite{b10}, researchers investigated the diversity in ensemble systems using an MCS composed of various techniques for feature representations, such as sparse and dense vector representations, and distinct learning algorithms, such as statistical, symbolic, ensemble and deep learning, and showed the advantage of adopting a heterogeneous and diverse pool of classifiers of MCS. In addition, \cite{pereira2025multiviewautoencodersfakenews} integrates the use of distinct feature representation techniques in a way to achieve a more consistent representation and demonstrates that the selected system of views to the final representation that feeds the algorithm is accurate in relation to only one representation technique. 

From both works, it is relevant to highlight that the exploitation of techniques from these two groups enriches the ensemble analysis in a search for diversity and is a successful approach to find it. Nevertheless, \cite{b10} makes a manual selection of a pool that requires visual selection and human efforts. Moreover, to the need for automating the selection of the pool, \cite{pereira2025multiviewautoencodersfakenews} shows that there is a more informative subset than the use of a combination of all feature representations, which is still an open problem. Our work addresses these limitations through an automated hierarchical selection approach that systematically identifies diverse classifier subsets while preserving the performance benefits established in prior work. 

\section{PROPOSED METHOD: HSFN}
\label{section:proposed_method}
\begin{figure*}[!ht]
    \centering
    \includegraphics[width=0.75\linewidth]{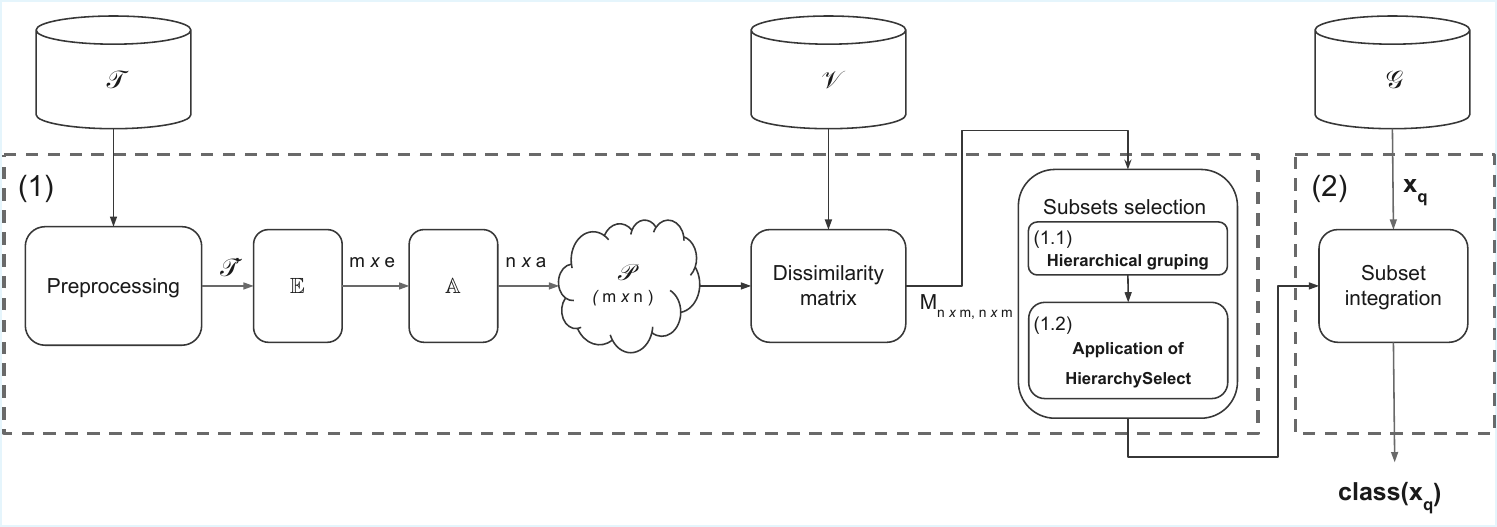}
    \caption{Illustration of HSFN. The method consists of 2 phases: (1) training and (2) testing. The $\mathcal{T}$ dataset corresponds to the training data, $\mathcal{V}$ to the validation data, and $\mathcal{G}$ to the testing data. $\mathbb{E}$ represents the feature representation and $\mathbb{A}$ the algorithm.  The training data is pre-processed and transformed into a numerical format by $m$ feature representation techniques and fed to the $n$ classification algorithms to obtain the pool of classifiers, represented by $P$.}
    \label{fig:proposed_method_methodology_2_fases}
\end{figure*}

The Hierarchical Selection for Fake News detection (HSFN) method constructs a diverse Multiple Classifier System (MCS) via hierarchical clustering. As illustrated in Figure~\ref{fig:proposed_method_methodology_2_fases}, HSFN operates in two phases: (1) training and (2) testing. In the training phase, the dataset $\mathcal{T}$ (training data) is used to generate a pool of classifiers $P$, which is then applied in the testing phase to predict labels for unseen data $\mathcal{G}$ (test data). HSFN's novelty lies in its hierarchical clustering-based selection mechanism, which systematically maximizes classifier diversity. The proposed heuristic, $HierarchySelect$, groups classifiers by dissimilarity and selects an optimal subset for MCS construction. Below, we detail each phase.

\subsection{Training phase}

In phase (1), we preprocess the text sentences in $\mathcal{T}$ and encode class labels numerically. Let $\mathbb{E} = \{E_1, \dots, E_m\}$ denote the set of feature extraction methods (e.g., TF-IDF, BERT embeddings) that convert text into numerical representations, and $\mathbb{A} = \{A_1, \dots, A_n\}$ the set of classification algorithms (e.g., SVM, Random Forest). For each pair $(E_i, A_j) \in \mathbb{E} \times \mathbb{A}$, we train a classifier that learns patterns from the features $E_i(\mathcal{T})$ and labels. This yields a pool $P$ of $m \times n$ classifiers. To mitigate overfitting (as in \cite{b10}), each classifier is evaluated on validation data. We then compute a dissimilarity matrix, where each entry captures the pairwise dissimilarity between two classifiers' predictions. Finally, we select a diverse subset from $P$ and integrate it into a final MCS for test-time prediction.

The automatic selection mechanism consists of two steps of the Subsets selection step: (1.1) hierarchical grouping and (1.2) application of the $HierarchySelect$ heuristic. In step (1.1), we apply hierarchical clustering to group classifiers based on their dissimilarity, producing a dendrogram. The clustering uses a linkage function (e.g., Complete, Single, Average, or Centroid~\cite{wierzchon2018modern}) to merge clusters iteratively. In step (1.2), $HierarchySelect$ dynamically cuts the dendrogram at varying hierarchy levels, extracting subsets of maximally diverse classifiers. Unlike \cite{b10}, which relies on manual selection of classifiers through visualization, our method automates this process by systematically evaluating diversity at each hierarchical level and selecting optimal ensembles without human intervention.

\begin{figure*}[htp]
\centering
\includegraphics[width=0.75\linewidth]{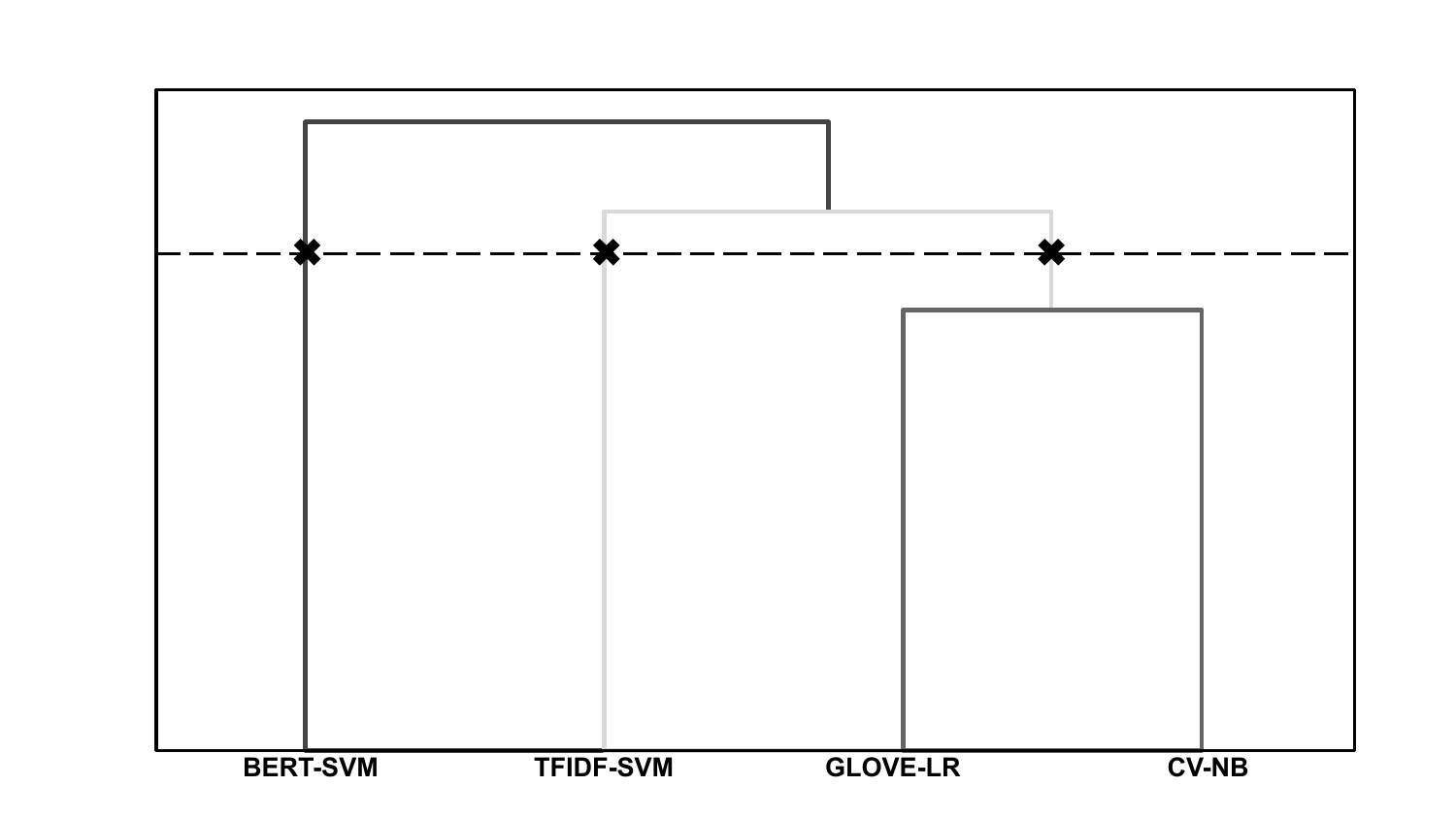}
\caption{Example of dendrogram with four classifiers. In this case, for HSFN subsets selection, the $k$ is 3 because the horizontal auxiliary line crosses 3 times the graph branches and finds 3 groups used to compose a final $P$ with 3 elements. The graph has a divisive agglomeration structure in a \textit{top-down} hierarchy. The X-axis corresponds to the classifiers and the Y-axis to the distance between them. The vertical proximity of the lines shows the similarity between the classifiers.}\label{fig:proprosed_method_dendograms}
\end{figure*}

The $HierarchySelect$ algorithm analyzes dendrogram levels to select classifier subsets. Figure \ref{fig:proprosed_method_dendograms} demonstrates this process with an auxiliary line traversing the hierarchy vertically, revealing three classifier groups: (1) BERT-SVM, (2) TFIDF-SVM, and (3) GLOVE-LR paired with CV-NB, clustered by prediction dissimilarity. From each group, we select the top-performing classifier based on evaluation metrics (e.g., choosing GLOVE-LR over CV-NB). The resulting ensemble combines complementary capabilities: BERT-SVM for contextual analysis, TFIDF-SVM for lexical patterns, and GLOVE-LR for short-text bias detection. This selection method ensures diversity while optimizing performance, with semantic differences between classifiers informing critical design choices like linkage functions, particularly valuable for distinguishing nuanced cases like satire versus malicious intent.

Algorithm~\ref{equation:pseudo-algorithm} formally describes $HierarchySelect$. The algorithm takes as input a dissimilarity matrix $M$ and performance metrics $\mathcal{V}'$ for all classifier pairs $(E_i, A_j)$. Unlike \cite{b10}, which requires manual threshold selection, our algorithm automatically evaluates all possible hierarchy levels ($k$ from 1 to $m \times n$). At each level, it: (1) partitions classifiers into $k$ clusters using hierarchical clustering, (2) selects the highest-performing classifier from each cluster based on $\mathcal{V}'$ metrics, and (3) adds these classifiers to the MCS candidate set. To (1), a function called f\_cluster is employed to identify flat clusters at a given level and to return the cluster containing the classifier. The function receives the number of clusters at the specified level and applies a criterion, such as distance, to group proximate elements within the same cluster. This exhaustive search ensures optimal diversity-accuracy trade-offs across all possible ensemble sizes, eliminating the subjectivity inherent in visual inspection approaches.
 
\begin{algorithm}
\caption{HierarchySelect} 
\label{alg:hierarchy_select}
\begin{algorithmic}[1] 
\Require $M$: dissimilarity matrix ($m \times n$)
\Require $\mathcal{V}'$: Set of classifier pairs with their performance metrics
\Ensure Set of $m \times n$ MCS configurations
\State $MCS \gets \emptyset$ 
\State $Z \gets \text{linkage}(M)$ \Comment{Compute hierarchical clustering}
\For {$(\text{metric}, \text{pair}) \in \mathcal{V}'$}
    \For {$k \gets 1$ \textbf{to} $m \times n$} 
        \State $\text{clusters} \gets \text{fcluster}(Z, k)$
        \For {$\text{group} \in \text{clusters}$}        
            \If{$\text{metric}(pair) = \max(\text{group})$}
                \State $MCS \gets MCS \cup \{\text{pair}\}$
            \EndIf
        \EndFor
    \EndFor
\EndFor 
\State \Return $MCS$
\end{algorithmic} 
\label{equation:pseudo-algorithm}
\end{algorithm}

\subsection{Testing phase}  
In phase (2), each classifier subset selected by $HierarchySelect$ is integrated into a final ensemble using stacking. The validation predictions from all classifiers in a subset $MCS \subseteq P$ are used to train a meta-classifier, which then generates the final predictions for the test data $\mathcal{T}$. This approach enables adaptive weighting of classifier contributions based on their validation performance, resulting in improved generalization compared to individual classifiers or static combination rules~\cite{duin2002combining}.

Therefore, based on these two phases, HSFN provides a final classifier that reflects the diversity obtained from the pool selected at the hierarchical level. In this context, we look to see if the HSFN performance is comparable to other heuristic methods or to monolithic classifiers' performance. These points will be further examined and discussed by analyzing the experimental results.

\section{EXPERIMENTAL SETUP} 
\label{sec:exp_setup}
\begin{table}[h]
    \centering
    \footnotesize
    \caption{Experimental setup.}
    \begin{tabular}{cc}
        \hline
        \\
        Experimental element & Items\\
        \\
        \hline
        \\
        $\mathbb{E}$ & CV, TF-IDF, W2V, GLOVE, FAST \\
        $\mathbb{A}$ & SVM, LR, RF, NB, MLP, ET, CNN, KNN \\
        Meta-classifier & LR, RF, NB \\
        Dataset & Liar, Senti, Covid, Fa-Kes, Ott, Kaggle \\
        \shortstack{Number of Labels} & 6 , 2\\
        Contexts & \shortstack{Politics, Health, Syrian War, \\ Tourism, Diverse topics}\\ 
        \\
        \hline
    \end{tabular}
    \\
    \label{tab:experimental_config} 
\end{table}

Table~\ref{tab:experimental_config} summarizes our experimental configuration. We selected five feature representations (EE) spanning different approaches: CountVectorizer (CV) and TF-IDF for traditional bag-of-words modeling, and Word2Vec (W2V), GloVe, and FastText for distributed word embeddings. These techniques were chosen based on their established effectiveness in fake news detection \cite{b12,b13,b19} and their ability to provide diverse feature spaces for classifier analysis. While transformer-based models like BERT offer superior contextual understanding, we excluded them from this study due to their substantial computational requirements for ensemble systems. The selected representations provide sufficient diversity for analyzing classifier behavior while maintaining practical computational efficiency.

For classification algorithms ($\mathbb{A}$), we implement eight approaches spanning all major paradigms: Support Vector Machines (SVM), Logistic Regression (LR), Random Forest (RF), Extra Trees (ET), Naïve Bayes (NB), Multi-Layer Perceptron (MLP), Convolutional Neural Networks (CNN), and K-Nearest Neighbors (KNN), following \cite{b10,b13,b14}. This creates 40 classifier variants through the full combination of representations and algorithms $|\mathbb{E}| \times |\mathbb{A}|$. Hyperparameters for traditional algorithms use scikit-learn defaults, while MLP and CNN architectures follow \cite{b10}, with output layers adapting to each dataset's class count (2-6 neurons). We evaluate performance using accuracy, precision, recall, and F1-score, consistent with \cite{b13,b14,b21,b24}. 

For classifier selection, we employ the double-fault diversity metric, which measures the proportion of instances where both classifiers in a pair make incorrect predictions. This choice is motivated by \cite{kuncheva2003measures}, which demonstrates that its inverse correlates with ensemble accuracy, making it particularly suitable for selecting complementary classifiers. The subsequent ensemble combination uses stacking implemented via DESlib \cite{b25}, with three alternative meta-classifiers: Logistic Regression (LR), Random Forest (RF), and Naïve Bayes (NB). These meta-classifiers were selected based on their proven effectiveness in similar ensemble frameworks \cite{b22,b23}.

We evaluate on six datasets used for the fake news detection,
chosen to cover varied contexts, for a more robust evaluation.
The Liar dataset \cite{b11} has six labels for political news. The Senti dataset \cite{b21} uses the same political context as Liar but with two labels. The Covid dataset \cite{b2} is about health. The Fa-kes dataset covers the Syrian war. The Ott dataset \cite{b24,b28} is about tourism. The Kaggle dataset \cite{b14} includes diverse topics. As in \cite{b45}, we preprocess the text by removing punctuation and contractions, fixing spelling, deleting URLs and IP addresses, making words lowercase, stemming, removing stopwords, and deleting words that appear only once \cite{b12}. For Fa-kes, Ott, and Kaggle datasets, similar to \cite{b14}.

\section{RESULTS}
\label{section:results}
\subsection{Comparison of Selection Strategies}
\label{subsection:benchmark_heuristics}

We evaluate HSFN against monolithic classifiers and three alternative selection heuristics from \cite{b10}: Group A (classifiers from a single algorithm), Group B (classifiers using a single representation technique), Group C (all monolithic classifiers combined), and our proposed Group D (HSFN-selected classifiers). Each group is identified by its letter designation, followed by the specific technique (for Groups A and B) and the meta-classifier employed. Table \ref{table: tabela} compares their performance across all datasets and metrics, where results for Fa-kes, Ott, and Kaggle represent 10-fold averages. The analysis prioritizes solutions that achieve comparable accuracy with fewer classifiers, with the number in parentheses indicating the classifier count for each approach.

\begin{table}[!ht]
    \centering
    \footnotesize
    \caption{Classifiers or groups of them that achieved best results for each metric and dataset. The number in parentheses means the number of classifiers in the group.}
    \begin{tabular}{ccccc}
    \hline
    \\
        Dataset  & Accuracy & Precision & Recall & F1-score \\
        \\
        \hline
        Liar & \shortstack{MLP-FAST} & SVM-W2V & C-RF (40) & NB-CV \\ 
        Senti & D-NB (1) & D-NB (1) & D-LR (5) & D-LR (5) \\ 
        Covid & KNN-CV & KNN-CV & CNN-GLOVE & C-RF (40) \\
        Fa-Kes & RF-TFIDF & RF-TFIDF & SVM-FAST, & B-CV-LR (8) \\
         &  &  & LR-FAST, &  \\
         &  &  & LR-W2V & \\
        Ott & NB-TFIDF & D-RF (20) & D-RF (10) & D-RF (20) \\ 
        Kaggle & C-RF (40) & C-RF (40) & EXTRA-TFIDF & C-RF (40) \\ \hline
    \end{tabular}
    \label{table: tabela}
\end{table}

\subsection{Comparison with the state-of-the-art}
\label{subsection:benchmark_literature}

We evaluate HSFN against three comparison approaches: (1) existing methods from the literature, (2) an Elbow selection method, and (3) a Baseline classifier. Following \cite{b27}, the Elbow method determines the optimal number of classifier groups by analyzing the balance between inter-group distance and intra-group cohesion. The Baseline represents random classification performance, calculated as 100\% divided by the number of classes, yielding 50\% accuracy for binary classification (2-class datasets) and approximately 16.7\% for six-class problems.
Table \ref{table: table_Comparative} presents the comparative results across all approaches

\begin{table}[!ht]
    \centering
    \footnotesize
    \caption{Results found in Literature and by the HSFN for each data dataset and metric accuracy.}
    \begin{tabular}{cccccc}
        \hline
        \\
        Dataset & Approach & Result  & Baseline & Elbow &HSFN \\
        \\
        \hline
          Liar  & Hybrid CNN \cite{b11} & \textbf{0,274} & 0,167 & 0,232 & 0,241\\
          Senti &  BERT-Base & \textbf{0,700} & 0,500 & 0,664 & 0,677\\
          & + CNN \cite{b21} &  &  &  &  \\ 
          Covid & CNN-LSTM \cite{b3} & 0,930 & 0,500 & 0,898 & \textbf{0,931} \\ 
          Fa-Kes & CNN-RNN \cite{b12} & \textbf{0,600} & 0,500 & 0,499 & 0,526\\ 
          &&&&(0,037) &  \\
          Ott & LIWC + Bigrams & \textbf{0,898} & 0,500 &  0,800 & 0,865 \\
          & + SVM \cite{b24} & &  & (0,033) &  \\ 
        Kaggle & RF \cite{b14} & 0,950 & 0,500 & 0,931 & \textbf{0,987} \\ 
        & & & &(0,026) & \\
        \hline
    \end{tabular}
\label{table: table_Comparative}
\end{table}

\subsection{Discussion}
\label{subsection:discussion}

The performance comparison in Table \ref{table: tabela} reveals that monolithic classifiers achieved the highest results in 45.8\% of cases, followed by the proposed method (29.2\%), Group C (20.8\%), and Group B (4.2\%). While Group A did not outperform any other approach, the proposed method demonstrated superior performance compared to Groups A-C, confirming the effectiveness of our selection heuristic. Notably, the proposed method accomplished these results using only 20 classifiers (50\% of all available monolithic classifiers), highlighting the value of selective diversity over exhaustive combinations. However, the results also show that certain monolithic classifiers can outperform all ensemble groups, including our proposed method, in specific contexts.

The proposed method's key advantage lies in its consistent generalization across diverse datasets while maintaining competitive performance with reduced computational requirements. This demonstrates an important trade-off in ensemble design: while individual classifiers may achieve peak performance in specific cases, systematic diversity selection yields more robust performance overall. The method's efficiency is particularly notable, as it explores only 40 possible combinations rather than the 1,099,511,627,775 potential combinations for MCS construction.

Comparative results with existing literature approaches (Table \ref{table: table_Comparative}) show that the proposed method consistently outperformed baseline implementations and achieved superior accuracy on the Covid and Kaggle datasets, while some literature methods achieved higher performance on other datasets, particularly \cite{b11, b21}, which incorporated additional non-text features for the Liar and Senti, respectively. These variations likely stem from differences in dataset characteristics, including size, content domain, and class distribution. Despite these factors, the proposed method demonstrates competitive performance across all evaluation scenarios, confirming its viability for practical fake news detection applications.

\section{CONCLUSION}
\label{section:conclusion}

This work introduced the method HSFN for selecting a diverse subset of classifiers for FN detection using hierarchical grouping and a double-fault diversity metric. By leveraging heterogeneous classifiers trained on multiple algorithms and feature representations and exploring the hierarchical levels, the approach improved classification performance while ensuring diversity. Experimental results on six datasets showed that the proposed method outperformed or matched existing selection heuristics, including baseline models and the Elbow approach.

These findings highlight the effectiveness of diversity-aware selection in MCS for FN detection, demonstrating its potential to enhance robustness across different datasets and classification tasks. Future work includes expanding the method to multilingual datasets, refining feature representations, and optimizing the integration step through multi-level stacking.


\begin{thebibliography}{99}


\bibitem{b1}{Allcott, Hunt, and Matthew Gentzkow. "Social media and fake news in the 2016 election." Journal of economic perspectives 31.2 (2017): 211-236.}

\bibitem{b2}{Patwa, Parth, et al. "Fighting an infodemic: Covid-19 fake news dataset." International Workshop on  Combating Online Hostile Posts in Regional Languages during Emergency Situation. Cham: Springer International Publishing, 2021.}

\bibitem{b3}{Surendran, Pranav, et al. "Covid-19 fake news detector using hybrid convolutional and Bi-lstm model." 2021 12th International Conference on Computing Communication and Networking Technologies (ICCCNT). IEEE, 2021.}


\bibitem{b7}{Ludmila, I. Combining pattern classifiers: methods and algorithms. Wiley, 2004.}

\bibitem{kuncheva2003measures}{Kuncheva, Ludmila I., and Christopher J. Whitaker. "Measures of diversity in classifier ensembles and their relationship with the ensemble accuracy." Machine learning 51.2 (2003): 181-207.}

\bibitem{b10}{Cruz, Rafael MO, Woshington V. de Sousa, and George DC Cavalcanti. "Selecting and combining complementary feature representations and classifiers for hate speech detection." Online Social Networks and Media 28 (2022): 100194.}

\bibitem{b11}{Wang, William Yang. "" liar, liar pants on fire": A new benchmark dataset for fake news detection." arXiv preprint arXiv:1705.00648 (2017).}

\bibitem{b12}{Nasir, Jamal Abdul, Osama Subhani Khan, and Iraklis Varlamis. "Fake news detection: A hybrid CNN-RNN based deep learning approach." International journal of information management data insights 1.1 (2021): 100007.}

\bibitem{b13}{Priya, Anu, and Abhinav Kumar. "Deep ensemble approach for COVID-19 fake news detection from social media." 2021 8th International Conference on Signal Processing and Integrated Networks (SPIN). IEEE, 2021.}

\bibitem{b14}{Ahmad, Iftikhar, et al. "Fake news detection using machine learning ensemble methods." Complexity 2020.1 (2020): 8885861.}

\bibitem{b15}{Hansrajh, Arvin, Timothy T. Adeliyi, and Jeanette Wing. "Detection of online fake news using blending ensemble learning." Scientific Programming 2021.1 (2021): 3434458.}

\bibitem{b19}{Taher, Youssef, Adelmoutalib Moussaoui, and Fouad Moussaoui. "Automatic fake news detection based on deep learning, FasTtext and news title." International Journal of Advanced Computer Science and Applications 13.1 (2022).}

\bibitem{b21}{Upadhayay, Bibek, and Vahid Behzadan. "Sentimental liar: Extended corpus and deep learning models for fake claim classification." 2020 IEEE International Conference on Intelligence and Security Informatics (ISI). IEEE, 2020.}

\bibitem{b22}{Cruz, Rafael MO, Robert Sabourin, and George DC Cavalcanti. "META-DES. H: A dynamic ensemble selection technique using meta-learning and a dynamic weighting approach." 2015 International Joint Conference on Neural Networks (IJCNN). IEEE, 2015.}

\bibitem{b23}{Boldrin, Fabiana Coutinho, et al. "Multi-Level Stacking." Encontro Nacional de Inteligência Artificial e Computacional (ENIAC). SBC, 2022.}

\bibitem{duin2002combining}{Duin, Robert PW. "The combining classifier: to train or not to train?." 2002 International conference on pattern recognition. Vol. 2. IEEE, 2002.}

\bibitem{mikolov2013efficient}{Mikolov, Tomas, et al. "Efficient estimation of word representations in vector space." arXiv preprint arXiv:1301.3781 (2013).}

\bibitem{pennington2014glove}{Pennington, Jeffrey, Richard Socher, and Christopher D. Manning. "Glove: Global vectors for word representation." Proceedings of the 2014 conference on empirical methods in natural language processing (EMNLP). 2014.}

\bibitem{devlin2019bert}{Devlin, Jacob, et al. "Bert: Pre-training of deep bidirectional transformers for language understanding." Proceedings of the 2019 conference of the North American chapter of the association for computational linguistics: human language technologies, volume 1 (long and short papers). 2019.}


\bibitem{pkekalska2002discussion}{{Pekalska}, Elzbieta, Robert PW Duin, and Marina Skurichina. "A discussion on the classifier projection space for classifier combining." International workshop on multiple classifier systems. Berlin, Heidelberg: Springer Berlin Heidelberg, 2002.}

\bibitem{farhangian2024fake}{Farhangian, Faramarz, Rafael MO Cruz, and George DC Cavalcanti. "Fake news detection: Taxonomy and comparative study." Information Fusion 103 (2024): 102140.}

\bibitem{wierzchon2018modern}{Wierzchoń, Sławomir T., and Mieczysław A. Kłopotek. Modern algorithms of cluster analysis. Springer, 2018.}

\bibitem{pereira2025multiviewautoencodersfakenews}{Pereira, Ingryd VST, George DC Cavalcanti, and Rafael MO Cruz. "Multi-view autoencoders for fake news detection." 2025 IEEE Symposium on Computational Intelligence in Natural Language Processing and Social Media (CI-NLPSoMe). IEEE, 2025.}

\bibitem{shu2017fake}{Shu, Kai, et al. "Fake news detection on social media: A data mining perspective." ACM SIGKDD explorations newsletter 19.1 (2017): 22-36.}

\bibitem{bontridder2021role}{Bontridder, Noémi, and Yves Poullet. "The role of artificial intelligence in disinformation." Data \& Policy 3 (2021): e32.}

\bibitem{b24}{Ott, Myle, et al. "Finding deceptive opinion spam by any stretch of the imagination." arXiv preprint arXiv:1107.4557 (2011).}

\bibitem{b25}{Cruz, Rafael MO, et al. "DESlib: A Dynamic ensemble selection library in Python." Journal of Machine Learning Research 21.8 (2020): 1-5.}

\bibitem{b27}{D’Silva, Jovi, and Uzzal Sharma. "Unsupervised automatic text summarization of Konkani texts using K-means with Elbow method." Int J Eng Res Technol 13.1 (2020): 2380.}

\bibitem{b28}{Ott, Myle, Claire Cardie, and Jeffrey T. Hancock. "Negative deceptive opinion spam." Proceedings of the 2013 conference of the north american chapter of the association for computational linguistics: human language technologies. 2013.}

\bibitem{b30}{Guo, Quanjiang, et al. "Tiefake: Title-text similarity and emotion-aware fake news detection." 2023 International Joint Conference on Neural Networks (IJCNN). IEEE, 2023.}

\bibitem{b31}{Dong, Diwen, et al. "Sentiment-aware fake news detection on social media with hypergraph attention networks." 2022 IEEE International Conference on Systems, Man, and Cybernetics (SMC). IEEE, 2022.}


\bibitem{b34}{Nirav Shah, Minal, and Amit Ganatra. "A systematic literature review and existing challenges toward fake news detection models." Social Network Analysis and Mining 12.1 (2022): 168.}


\bibitem{b37}{Garrido-Merchan, Eduardo C., Roberto Gozalo-Brizuela, and Santiago Gonzalez-Carvajal. "Comparing BERT against traditional machine learning models in text classification." Journal of Computational and Cognitive Engineering 2.4 (2023): 352-356.}


\bibitem{b41}{Alsuwat, Emad, and Hatim Alsuwat. "An improved multi-modal framework for fake news detection using NLP and Bi-LSTM." The Journal of Supercomputing 81.1 (2025): 177.}

\bibitem{b44}{Cruz, Rafael MO, et al. "Feature representation selection based on classifier projection space and oracle analysis." Expert Systems with Applications 40.9 (2013): 3813-3827.}

\bibitem{b45}{Khan, Junaed Younus, et al. "A benchmark study of machine learning models for online fake news detection." Machine Learning with Applications 4 (2021): 100032.}


\end{thebibliography}
\end{document}